\documentclass[fleqn,10pt]{wlscirep}
\usepackage[utf8]{inputenc}
\usepackage[T1]{fontenc}
\usepackage{subfig}
\usepackage{graphicx}
\usepackage{multirow}
\usepackage{multicol}
\usepackage{xcolor}
\usepackage{soul}
\usepackage{algorithmic}
\usepackage[ruled,linesnumbered,lined]{algorithm2e}
\usepackage{amssymb}
\usepackage{trimclip}
\mathchardef\mhyphen="2D

\def\AR{\clipbox{0pt 0pt .35em 0pt}{\textit{\bfseries A}}\kern-.05emR}

\title{Counting Manatee Aggregations using Deep Neural Networks and Anisotropic Gaussian Kernel}

\author[1]{Zhiqiang Wang}
\author[1]{Yiran Pang}
\author[1]{Cihan Ulus}
\author[1,*]{Xingquan Zhu}
\affil[1]{Department of Electrical Engineering and Computer Science, Florida Atlantic University, Boca Raton, FL 33431, USA. Email: \{zwang2022, ypang2022, culus2021, xzhu3\}@fau.edu}

\affil[*]{corresponding author}

\begin{abstract}
Manatees are aquatic mammals with voracious appetites. They rely on sea grass as the main food source, and often spend up to eight hours a day grazing. They move slow and frequently stay in group (\textit{i.e.} aggregations) in shallow water to search for food, making them vulnerable to environment change and other risks. Accurate counting manatee aggregations within a region is not only biologically meaningful in observing their habit, but also crucial for designing safety rules for human boaters, divers, \textit{etc.}, as well as scheduling nursing, intervention, and other plans. In this paper, we propose a deep learning based crowd counting approach to automatically count number of manatees within a region, by using low quality images as input. Because manatees have unique shape and they often stay in shallow water in groups, water surface reflection, occlusion, camouflage \textit{etc.} making it difficult to accurately count manatee numbers. To address the challenges, we propose to use Anisotropic Gaussian Kernel (AGK), with tunable rotation and variances, to ensure that density functions can maximally capture shapes of individual manatees in different aggregations. After that, we apply AGK kernel to different types of deep neural networks primarily designed for crowd counting, including VGG, SANet, Congested Scene Recognition network (CSRNet), MARUNet \textit{etc.} to learn manatee densities and calculate number of manatees in the scene. By using generic low quality images extracted from surveillance videos, our experiment results and comparison show that AGK kernel based manatee counting achieves minimum Mean Absolute Error (MAE) and Root Mean Square Error (RMSE). The proposed method works particularly well for counting manatee aggregations in environments with complex background. 

\end{abstract}
\begin{document}
\flushbottom
\maketitle

\section*{Introduction}

Recent advancements in Artificial Intelligence (AI) have allowed it to be used in a wide spectrum of fields~\cite{zhu2007knowledge}. Especially, AI applications in ecology to save species have gained more importance in the last decade with the increasing number of endangered animals. There are many ways to use computational methods to help save endangered species, such as detection the presence of the species~\cite{corcoran2019automated} and counting species to collect information about numbers and density \textit{etc.} In this paper, we are proposing a deep neural network based method to help save manatees, an endangered species.

Human activities impact environment in numerous ways including deforestation, overpopulation, poaching, over-fishing, and climate change. These negative impacts directly affect the physical environment, posing risks and generating opportunities for wildlife. Sometime, this dispersion benefits species, mostly invasive species~\cite{banks2015role}. In most cases, however, human-assisted dispersion adversely affects the natural world. As a result, wildlife populations are declining at an alarming rate, and the extinction rates are now up to 100 times higher than the normal extinction rate~\cite{ceballos2015accelerated}. 

Manatees are one of the wildlife species being affected by human-related threads~\cite{quintana2010regional}. There are four species of manatees in the world: Trichechus inunguis (Amazonian manatee), Trichechus pygmaeus (dwarf manatee), Trichechus senegalensis (West African manatee) and Trichechus manatus (West Indian manatee)~\cite{henaut2022cognition}. Amazonian manatees and dwarf manatees populate in freshwater habitats. The dwarf manatees are contentious species only found in the river Aripuaña in Brasil and closely linked to the Amazonian manatee~\cite{van2015hotspot, vianna2006phylogeography}. The West African manatees  are distributed from Angola to Senegal~\cite{vianna2006phylogeography}. The West Indian manatee prefers shallow coastal habitats such as rivers and estuaries. They can be found from Brazil to Florida and all the way around the Caribbean islands. The West Indian Manatee has two subspecies; Florida Manatee and Antillean Manatee, and both of them are considered endangered by IUCN (International Union for Conservation of Nature) ~\cite{deutsch2008trichechus}. 

While some manatee species, such as the Florida Manatee, have a relatively limited geographical distribution, some manatee types inhabit a wide geographical range. As a result of this wide range, individual manatee migrations occur~\cite{nourisson2011evidence}. For example, during the 2021 winter, a manatee migration from Florida to the Mexican coast was observed~\cite{castelblanco2021first}. Furthermore, some types of manatees, such as Antillean manatees, can live in a diverse range of ecosystems. In some cases, they inhabit clear saltwater in Belize and Mexico, visit rivers, reefs, and freshwater lagoons, whereas they live in salt or fresh water in Chiapas and Tabasco states with insufficient visibility~\cite{corona2021searching,rodas2008distribution}. Such diverse living habits and behaviors make it inherently difficult to track them. 

Over the last decades, manatee populations have been continuously decreasing. Manatees tend to live as a group or individuals. Most of their populations exists as tiny isolates and are also low in density~\cite{keith2015trichechus}. Furthermore, they are frequently scattered throughout huge bodies of water and display evasive behavior due to hunting pressure, making detection and counting challenging with existing approaches. Knowing the number of manatees and their gathering pattern in real-time is vital for understanding their population dynamics. The timeliness and accuracy of the count data upon which choices are made frequently determine the efficacy of management decision-making. In other words, improvements in counting techniques may portend better ecological results from management decisions. Meanwhile, manatees rely on sea grass as their primary food source. Because sea grass requires sunlight and shallow water to grow, manatees tend to stay in shallow water to hunt food, making them vulnerable to the environment, \textit{e.g.} they have very little room/time to move away from (avoid) oncoming boats, resulting in deadly collisions if the boat drivers are not aware that they are approaching manatees.

Using aerial survey data, counting estimates for manatees in southern Florida, USA, was developed, and environmental and temporal factors were discovered to impact distributions~\cite{edwards2021monitoring,bauduin2013index}. However, aerial surveys are time-consuming and costly, and the accuracy depends on factors such as observer bias, weather, and time of the day. Consequently, less time-consuming and less costly counting methods gain more importance in detecting the number of  manatees. Furthermore, it is also crucial to have a method to provide a real-time count to allow ecologists to be aware of the threat early and act proactively to protect manatees.

Although there are few researches applied to wildlife counting~\cite{arteta2016counting}, most of the existing counting methods~\cite{zhang2016single,zhang2015cross,pang2023federated} are applied to crowds to count the number of people, due to their relevance to important applications such as urban planning and public safety. Fortunately, such advanced techniques in crowd counting can also be generalized to other fields such as wildlife counting, by taking specific characteristics of the objects into consideration. 

In this paper,  we propose to use crowd counting methods to count manatee aggregations. Our goal is to accurately estimate number of manatees in a specific region, using low quality images as input. Due to numerous factors, as we have elaborated above, manatee counting is a challenging task. 
\begin{itemize}
\item{Occlusion:} Because manatees tend to live in herd. They frequently block each others when viewing from the surface. 
As a result, small manatees are likely to be partially or completely blocked from the view. 
\item{Distributions and Distortion:} Due to diverse living habits and behaviors, manatees often present in different population density, perspective distortions, and lightning conditions. Without sufficient training data for each type of scene, it is difficult for a model to obtain accurate results for counting. 
\item{Reflections and Camouflage:} Furthermore, water reflections tend to make manatees invisible in reflection areas, counting manatees from images captured from surface mounted camera is difficult. 
\item{Background:} Finally, the high similarity of appearance between manatees and some elements in the background, such as fishes, rocks, imposes additional challenge to manatee counting. 
\end{itemize}
In order to address the above challenges and accurately estimate the density of the manatee, we propose a deep neural network based crowd counting method, which learn to estimate manatee density within an input image. Our method considers distortions caused by the perspective between the water space and the image plane. Furthermore, since the shape of the manatee is closer to an ellipse than a circle, we propose a method that uses an Anisotropic Gaussian kernel (AGK) to best represent the manatee contour, and estimate manatee density in the scene. By formatting manatee counting as a deep neural network density estimation learning task, our approach balances the labeling costs \textit{vs.} counting efficiency. As a result, our method delivers a simple and high throughput solution for manatee counting requiring very little labeling efforts.

\subsection*{{Contribution}}
Our research brings the following three unique contributions to enrich our data and algorithm design for domain specific tasks:
\begin{itemize}
\item{Deep Learning for Counting Manatees:} We are among the first to introduce deep learning method to automatically count manatee through low-resolution images captured from surface mounted camera. This pioneering study not only addresses the technical challenges of counting in complex outdoor environments but also offers potential ways to aid endangered species.

\item{Anisotropic Gaussian Kernel with Line Label Annotation:} We introduce Anisotropic Gaussian Kernel combined with line label annotations to generate density map. This novel method can represent the unique shapes of manatees and deliver more precise counting results.

\item{Manatee Counting Dataset:} To validate our method and facilitate further research in this domain, we have developed a comprehensive manatee counting dataset, published through Github for public access (https://github.com/yeyimilk/deep-learning-for-manatee-counting).
\end{itemize}

\section*{Related Work}
Our research is closely related to two research tasks: (1) learning to count number of objects within a scene; and (2) obtaining labels to support counting.

\subsection*{Counting Methods}
Given the above-mentioned specific significance of counting, an increasing number of scholars have attempted to address the counting issue. Existing methods in the field mainly fall into three groups: detection-based~\cite{enzweiler2008monocular,leibe2005pedestrian,wu2005detection}, regression-based, and density estimation based approaches including Convolutional Neural Network (CNN) based density estimation techniques~\cite{gao2020cnn}. It is worth noting that with the superb performance of deep learning methods, models based on CNN have largely dominated a variety of counting tasks.

Previous counting methods mainly focused on counting people, although some methods focused on cars, animals, cells etc. Early works~\cite{leibe2005pedestrian,wu2005detection,topkaya2014counting}  focused on counting people, using detection-based approach. Detection-based methods usually detect a head or person by using a sliding-window-like detector to count the number of people~\cite{dollar2011pedestrian}. Recently many object detection tools (YOLO~\cite{redmon2016you}, R-CNN~\cite{ren2015faster} etc.) are developed for object detection in sparse scenes. 
Nevertheless, these approaches do not show good performance in congested scenes because they require extraction of low-level characteristics.

Crowd counting, especially in real complex environments, has received increasing attention in various domains. A recent survey\cite{farjon2023deep} indicates that majority count-related studies predominantly center around yield estimation, phenotyping, livestock monitoring, and insect monitoring, which together constitute approximately 97\% of the applications\cite{farjon2023deep}.

A modified version of the Inception-ResNet architecture was employed to count tomatoes and simulated synthetic images were also used to enhance accuracy, although the system struggles to count green fruits\cite{rahnemoonfar2017deep}. A novel rice plant counting network, termed RPNet\cite{bai2023rpnet}, has four modules: feature encoder, attention block, initial density map generator, and attention map generator. Results indicate that RPNet outperforms MCNN, CSRNet, SANet, TasselNetV2, and FIDTM on certain high-resolution image datasets. Many detection-based and segmentation-based network have been proposed for counting fruits or plants, including those specific to blueberries, wheat spikes, panicles, pistachios and grapes clusters\cite{farjon2023deep}.

The application of deep learning models, particularly CNN combined with aerial imagery captured through UAV (Unmanned Aerial Vehicle)\cite{lin2019transfer} has shown significant promise in detection and counting. Models such as NasNet, Xception, YOLOv4, and YOLOv5 have demonstrated high accuracy, even under challenging conditions. While these models succeed in detecting cattle in various conditions, challenges like occlusion and diverse cattle breeds remain areas for improvement\cite{de2023counting,barbedo2019study,rahnemoonfar2019discountnet}.

Our method addresses the unique challenges posed by counting submerged animals in outdoor open water environment, especially when dealing with long-shape entities such as manatees, from overhead webcam low quality images.

\begin{figure*}[btp]
  \centering
  \includegraphics[width=0.95\textwidth]{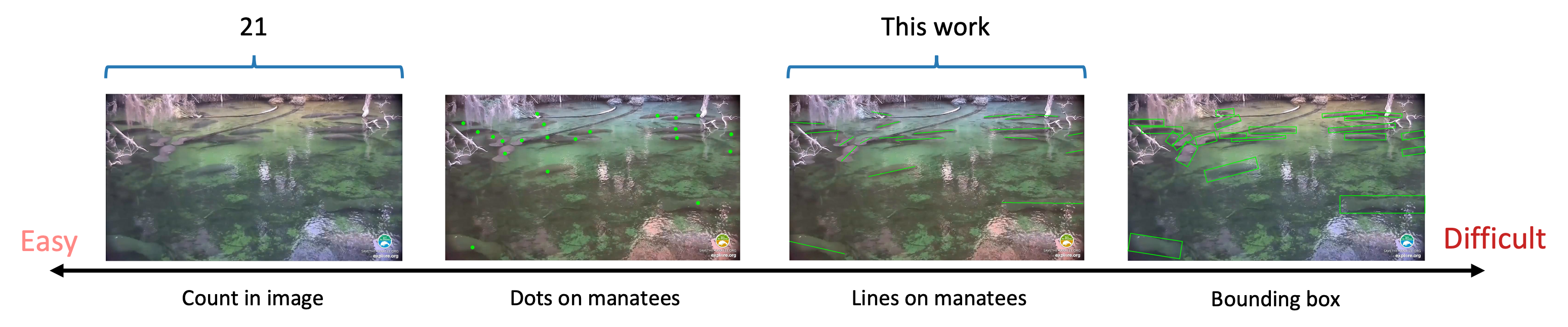}
    \caption{Comparison of different types of labels with their difficulties for the labeling work.}
    \label{fig:easy_to_difficult_comparison}
\end{figure*}

\subsection*{Annotation and Labeling for Counting}
Labeling is a critical step to support learning. In order to learn to estimate number of objects in a scene, it is necessary to provide supervision (\textit{i.e.} label information) for the learning algorithms. Such label information varies from labeling the whole image to each single object in the image. 

\subsubsection*{Total Count Annotation}
Using total count as a label in counting tasks can reduce annotation costs. This type of annotation is often referred to as Weakly-Supervised.
Earlier counting frameworks~\cite{chan2009bayesian,pham2015count} mostly used this approach. Borstel et al.~\cite{borstel2016gaussian} proposed a weakly-supervised solution based on the Gaussian process for density estimation. The training samples are divided into multiple subregions in their work, but each region is still annotated with only count labels.
Modern counting methods try to dig for more information. Yang et al.~\cite{yang2020weakly} used a soft-label sorting network to sort multiple images in the presence of only numerical labels. Then they used a regression network with a shared backbone to obtain the final number of people.
MATT~\cite{lei2021towards} and Sam et al.~\cite{sam2019almost} introduced a small number of point-level annotations and trained mainly using total count annotations. Their work shows that only a small number of point annotation samples can improve counting accuracy.
JCTNet~\cite{wang2022joint} and TransCrowd~\cite{liang2022transcrowd} use performant Transformer structures to regress the overall number directly. Further, CrowdMLP~\cite{wang2022crowdmlp} uses a more concise multilayer perceptron (MLP) as the overall architecture and introduces a self-supervised agent task to impose spatial cues implicitly.
Although the total count annotation-based approach has the lowest annotation cost, due to the lack of spatial information, the average counting performance is lower than that of the density map-based approach.

\subsubsection*{Bounding Box Annotation}
In order to provide accurate information of the object in the scene, bounding box annotations draws a minimum outer rectangle for each object appearing in the image. This allows underlying learning algorithms to precisely learn properties of the objects. Nevertheless, drawing precise rectangles is difficult and labor-intensive, especially for images with a high density of objects. In addition, bounding Box-based annotation methods cannot provide accurate predictions in extreme situations, such as low resolution or severe occlusion.

\subsubsection*{Dot Annotation}
Dot annotation, one of the most commonly used techniques in recent years, marks each counted object as a dot, where the dot is usually located at the center of the object, \textit{e.g.} the center of a person's head. As a result, dot annotation creates a locality point map where the sum of the point map equals to the total number of objects, and the learning can be carried out using information from the dots.

Dot annotation tremendously reduces the labeling efforts, compared to others like bounding box annotation, and is also able to handle extreme cases like object occlusions. In reality, it is, however, difficult to train a neural network (NN) using only sparse points. Therefore, point annotations are usually transformed into density maps, using transformation process defined as follows.
\begin{equation}
    \mathcal{D}_I = \sum_{x\in L\in I}\mathbf{K}_{\sigma}(x)
    \label{equ:base_network}
\end{equation}
Where $\mathbf{K}_{\sigma}(\cdot)$ is a kernel function (\textit{e.g.} a Gaussian kernel) and $\mathcal{D}_I$ is the density map of image $I$ (generated using kernel functions). $L$ is the set of labeled dot/point locations in the input image $I$, and $\sigma$ is the scale parameter of the 2-D Gaussian kernel. When integrated, the density map contains per-pixel density information of the scene, which results in the count of object in the image.

Intuitively, the transformation in Eq.~(\ref{equ:base_network}) blurs each point annotation according to the scale parameter $\sigma$, and settings of $\sigma$ represent different types of density mapping. The most basic way is to keep $\sigma$ as a fixed value, and an existing work~\cite{li2018csrnet} has studied commonly used $\sigma$ parameters.

One limitation of using a fixed $\sigma$ value is that it prevents the density map label from capturing perspective information of the counted objects. When the scales of the objects varies significantly in the image, a fixed $\sigma$ value results in low accuracy. Alternatively, an adaptive kernel~\cite{zhang2015cross} uses adaptive $\sigma$ values by taking average distance obtained from $k$-nearest neighbor algorithm into consideration. Intuitively, it produces a severely blurred density map in a highly dispersed regions of the target to accommodate the scale variation due to perspective. However, this approach relies on the uniform distribution of the counted objects. When discrete targets are present, the $k$-nearest algorithm becomes unreliable.

Modern approaches try to introduce additional information to produce more reliable adaptive kernels. Liu et al.~\cite{liu2019context}, Shi et al.~\cite{shi2019revisiting}, Yan et al.~\cite{yan2019perspective}, and Zhang et al.~\cite{zhang2015cross} use a perspective map to smooth the final density map so that larger counting targets close to the camera have larger and smoother Gaussian regions in an attempt to eliminate the error due to perspective distortion. However, perspective mapping is unavailable in manatee counting because the water is not on the same plane. Some work~\cite{lian2019density} uses depth camera to obtain relative depth of the target to the camera to estimate the size of the target. However, depth acquisition of underwater objects is still an open problem, and is uncommon (due to cost) in general surveillance/tracking systems.

Figure~\ref{fig:easy_to_difficult_comparison} shows an example of different labeling/annotation approaches on an image with 21 manatees in the scene. From left to right, the labeling costs and difficulty increase. The left most image is labelled with a number 21, which is the number of manatees in the scene (the label does not provide any additional information about manatees, such as locations, orientation \textit{etc.}). Dot annotation provides circular points to mark each manatees, and bounding-box annotation uses a rectangle box to outline region of each manatee. For comparisons, our proposed work is based on line-segment annotation where a single line-segment is used to label/mark each manatee.

\section*{Proposed Method}
Recently, density maps have been commonly used for presenting crowd counting because they can represent the distribution of the crowd. In our research, we propose to use Anisographic Gaussian Kernel (AGK) based crowd counting approach for manatee counting. In the following, we will first introduce kernel density based counting, and then propose manatee customized crowd counting framework.

\subsection*{Base Network}
Following recent approaches, we perform counting based on the density estimation framework. The input of the framework is an image $I\in\mathbb{R}^{w \times h}$, represented as a $w\times h$ matrix where $w$ and $h$ denote image width and height respectively. The ground-truth density map $\mathcal{D}_I\in\mathbb{R}^{w\times h}$ is used to train deep neural networks (DNNs) by imposing normalized 2-D Gaussian at manatee locations provided by the annotations. The deep neural network is trained to predict the density map $\hat{\mathcal{D}}_I\in\mathbb{R}^{w\times h}$ for an image $I$, such that the network predicted density map output $\hat{\mathcal{D}}_I$ is sufficiently close to the ground-truth $\mathcal{D}_I$.


\subsubsection*{Kernel Function} 
A kernel, commonly used in machine learning to perform classification and clustering, is a non-linear mapping of two vectors in a feature space, through the dot product of two vectors. Given two $d-$dimensional vectors $\mathbf{x}_{i},\mathbf{x}_{j}\in \mathbb{R}^d$, and a transformation function $\varphi(\mathbf{x})$ defined for each vector, a kernel mapping between $\mathbf{x}_{i}$ and $\mathbf{x}_{j}$ is a function defined as dot product $\varphi(\mathbf{x}_i)$ and $\varphi(\mathbf{x}_j)$ as follows:
\begin{equation}
    \label{fun:kernel1}
\mathbf{K}(\mathbf{x}_i,\mathbf{x}_j)={\varphi(\mathbf{x}_i)}^T\varphi(\mathbf{x}_j)
\end{equation}
For example, for $2-$dimensional vectors $\mathbf{x}_{i},\mathbf{x}_{j}\in \mathbb{R}^2$, with $\mathbf{x}_i=[x_{i,1},x_{i,2}]$ and $\mathbf{x}_j=[x_{j,1},x_{j,2}]$ a simple polynomial kernel is defined as
\begin{equation}
    \label{fun:kernel2}
\begin{split}
\mathbf{K}(\mathbf{x}_i,\mathbf{x}_j)& = (1+{\mathbf{x}_i}^T\mathbf{x}_j)^2 \\
&=(1+x_{i,1}x_{j,1}+x_{i,2}x_{j,2})^2\\
&=1+x^2_{i,1}x^2_{j,1}+x^2_{i,2}x^2_{j,2}\\
&+2x_{i,1}x_{j,1}+2x_{i,2}x_{j,2}+2x_{i,1}x_{i,2}x_{j,1}x_{j,2}\\
&={\varphi(\mathbf{x}_i)}^T\varphi(\mathbf{x}_j)
\end{split}
\end{equation}
where $\varphi(\mathbf{x}_i)=[1,x^2_{i,1},x^2_{i,2},\sqrt{2}x_{i,1},\sqrt{2}x_{i,2},\sqrt{2}x_{i,1}x_{i,2}]$, and $\varphi(\mathbf{x}_j)$ is defined similarly.

The kernel function should satisfy the following three properties: symmetrical, non-negative, and the area under the curve of the function must be equal to 1. 

There are some well-known examples of kernels satisfying specific properties, such as Gaussian kernel, multivariate Student kernel, and Laplacian kernel. The Gaussian kernel can be expressed as
\begin{equation}
    \label{gk}
\textbf{K}_{\sigma}(\mathbf{x}_i,\mathbf{x}_j)=(\frac{1}{\sqrt[]{2\pi}\sigma})^{d}\text{exp}(-\frac{\left\| \mathbf{x}_i -\mathbf{x}_j\right\|^{2}}{2\sigma^{2}})
\end{equation}

Similar to polynomial kernel, Gaussian kernel function in Eq.~(\ref{gk}) can be transformed as dot product between two vectors, $\varphi(\mathbf{x}_i)$ and $\varphi(\mathbf{x}_j)$, by using Taylor series to expand the kernel function into an infinite series of products.

The original Gaussian kernel has the same spread ($\sigma$) for all feature dimensions. As a result, it is difficult to represent a high dimensional space and differentiate features more important to capture the decisions (or classifications)~\cite{1417501}. 

\subsubsection*{Kernel Density Map Generation} Given an image $\mathbf{I}\in\mathbb{R}^{w\times h}$ represented as a $w \times h$ array, and a set of $n$ labelled points $\mathbf{p}_1,\cdots,\mathbf{p}_n$ in the image (\textit{e.g.}  using dot annotation), the kernel density map intends to generate a density map $\mathcal{D}_I\in\mathbb{R}^{w\times h}$ of $I$, with respect to the labelled points $L$ (so the density map is primarily focused on labelled points). 

The kernel density estimator (KDE) is a non-parametric estimator used to estimate the univariate or multivariate densities based on kernels as weights\cite{kim2012robust}. One direct way to create a kernel density map $\mathcal{D}_I$ is to compare each pixel in $I$ 
 to each labeled point $\mathbf{p}_i\in\mathbf{R}^2$, using 2-D location to represent pixels and points. 

Denote $\mathbf{x}_{i,j}\in\mathbb{R}^2$ the 2-D location of a pixel of image $I$ located at $[i,j]$, kernel density map $\mathcal{D}_I$ can be calculated using kernel transformation below
\begin{equation}
    \label{kde}
\mathcal{D}_I[i,j]=\frac{1}{n}\sum_{k=1}^{n}\textbf{K}_{\sigma}(\textbf{x}_{i,j},\textbf{p}_{k});
\end{equation}

where $i=1,\cdots,w; j=1,\cdots,h$ are indices of image width and height, and $\textbf{K}_{\sigma}$ is a kernel function with bandwidth $\sigma$. 

\subsubsection*{Density Map to Counting}
After obtaining the density map of an image, the number of manatees can be calculated by using element-wise summation of all points' density value as follows
\begin{equation}
    \begin{split}
        C_I = \sum^w_{i=1}\sum^h_{j=1}\mathcal{D}_I[i,j]
    \end{split}
    \label{equ:counting}
\end{equation}
where $C_I$ denotes ground-truth manatee numbers in the image $I$. For each labeled image, the annotations have ground-truth $C_I$ value, so density map needs to be normalized accordingly to ensure that the sum of element-wise density map equals to its $C_I$ value. 

\begin{figure*}[bth]
  \centering
  \begin{small}
  \includegraphics[width=0.8\textwidth]{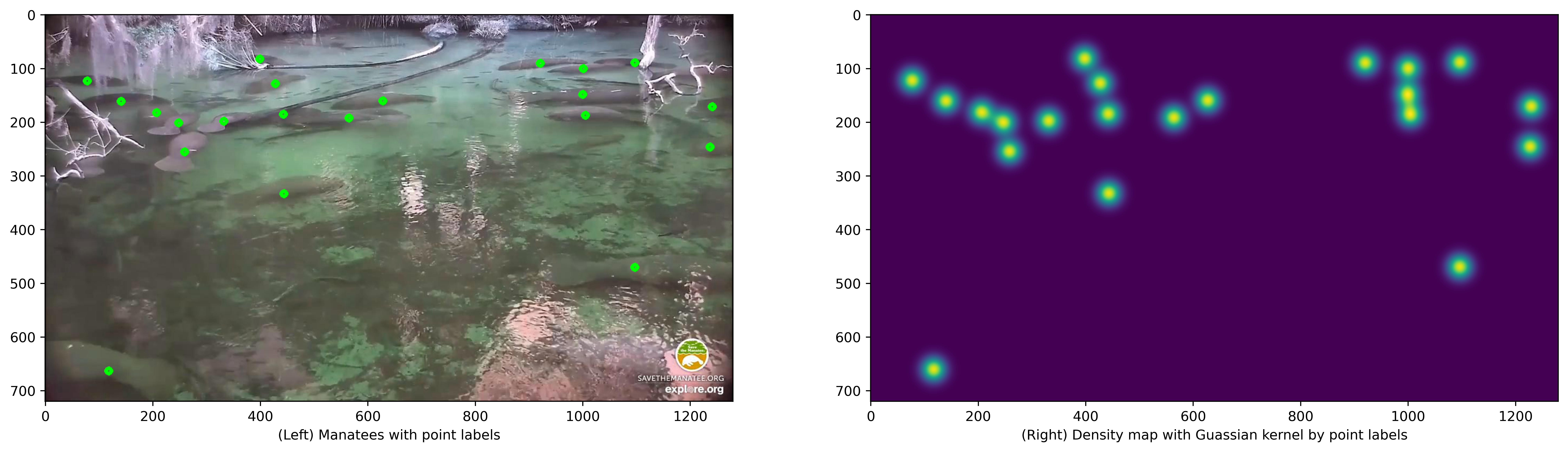}
    \caption{\emph{Left panel:} An image with dot labels of manatees, and \emph{Right panel:} Density map of the image generated by applying Gaussian distributions to labeled points}
    \label{fig:point_example}
    \end{small}
\end{figure*}

Figure ~\ref{fig:point_example} shows an example of an input image $I$ with dot annotations (left panel), and its density map $\mathcal{D}_I$ based on labeled points (right panel). The ground-truth $C_I$ is 21, because there are 21 manatees within the image. 

After obtaining the density map $\mathcal{D}_I$, we will train neural network using input image $I$, and using density map $\mathcal{D}_I$ as expected output of the network. A sufficiently trained neural network is therefore capable of learn to detect an input image's manatee locations as a density map. By calculating element-wise sum of the predicted density map, using Eq.~(\ref{equ:counting}), we can calculate number of manatees in the scene (detailed in the latter section).  

\subsection*{Manatee Customized Crowd Counting}
\subsubsection*{Line-Label for Manatee Counting}
Although point label are commonly used for generating density map in crowd counting, a manatee's body is oval-shaped, meaning that point annotations cannot effectively capture orientation and shape of manatees for accurate counting. An alternative solutions is to use line-segment labels, \textit{e.g.}, a straight line segment, to mark each manatee. 

A challenge of using a line-segment to annotate each manatee is that a line consists of an infinite number of points, making it difficult to obtain a density map of the input image. In this paper, we propose to use a limited number of points to represent the segment, and then use Gaussian kernel to denote each point for density map generation. 

More specifically, given two endpoints, $\mathbf{x}_a=\lbrack x_{a,x}, x_{a,y}\rbrack$ and $\mathbf{x}_b=\lbrack x_{b,x}, x_{b,y}\rbrack$, of a line-segment, we generate $\lceil |x_{b,x} - x_{a,x}| + 1\rceil$ number of Gaussian kernels using position evenly distributed on the line (where $\lceil \cdot \rceil$ denotes a ceiling function). In addition, to better capture manatee shapes, we generate an oval shaped Gaussian kernel for each dot, using $\sigma$ value, which is adjusted according to the position of current point $\mathbf{x}_i=\lbrack x_{i,x}, x_{i,y}\rbrack$ comparing to the two endpoints of the line segment, as defined in Eq.~(\ref{equ:line_scale}), where $a$ is a constant value (\textit{i.e.} a parameter).
\begin{equation}
    \sigma = \sigma_{basic} + a \cdot \min(\left\|\mathbf{x}_i - \mathbf{x}_a\right\|, \left\|\mathbf{x}_i - \mathbf{x}_b\right\|) 
    \label{equ:line_scale}
\end{equation}
After obtaining $\sigma$ value, we can use point $\mathbf{x}_i$'s current location as mean ($\mu$) and generate a Gaussian distribution corresponding to point $\mathbf{x}_i$. Repeating this process, one can generate many Gaussian kernels for each line-segment. The total number of Gaussian kernels for each input image (which often contains many line segments) are used to normalize Gaussian kernels such that the density map of the whole image follows a distribution. 

\begin{figure*}[ht]
  \centering
  \includegraphics[width=1\textwidth]{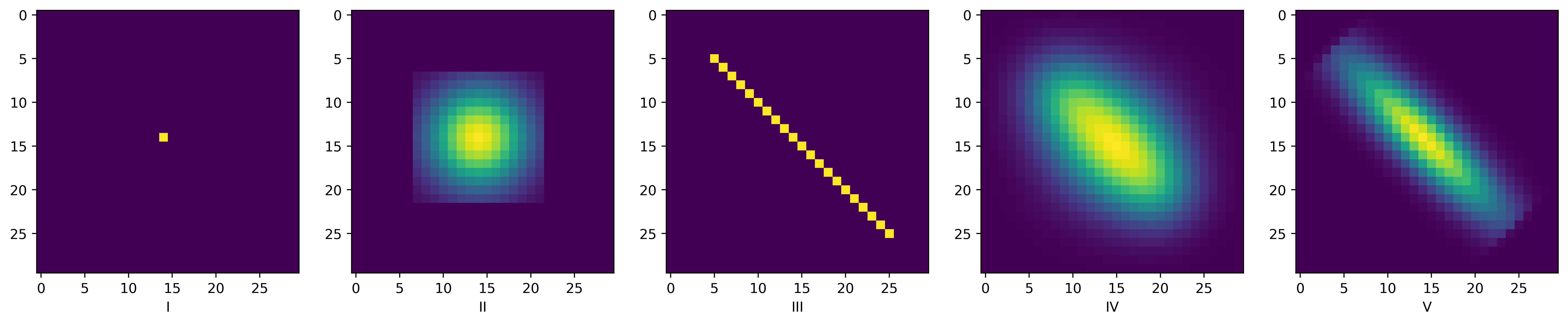}
    \caption{\emph{(I)} A single point at $(14, 14)$; \emph{(II)} Gaussian with $\sigma=4$; \emph{(III)} line-segment from point $(5, 5)$ to $(25, 25)$; \emph{(IV)} Gaussian with $\sigma_{basic}=3$ and $a=0.2$; \emph{(V)} Anisotropic Gaussian with $\sigma_{x_1}=8.24, \sigma_{x_2}=2.06$ for line-segment in \emph{(III)}}
    \label{fig:label_examples}
\end{figure*}

To illustrate the density map generation using line-segment, Figure~\ref{fig:label_examples} shows examples of using single point and a line-segment to generate manatee customized density map. From left to right, Figure~\ref{fig:label_examples}(I) denotes a single point at $(14,14)$, while the subfigure (II) is the density map generated by given head coordination of the Gaussian kernel with $\sigma=4$. Figure~\ref{fig:label_examples}(III) is a line-segment from point $(5,5)$ to point $(25,25)$ which includes 21 points in total, and Figure~\ref{fig:label_examples}(IV) denotes the density map generated by the given points in Figure~\ref{fig:label_examples}(III) where the result has been normalized. In this subfigure, the smallest $\sigma$, $\sigma_{basic} = 3$, is 3 at the two endpoints,  while the largest $\sigma$ value is 5 corresponding to the Gaussian kernel at the center point of the line-segment at $(15, 15)$. 

The results in Figure~\ref{fig:label_examples} show that using a line-segment, combined with Gaussian kernels, can provide a customized density map resembling a manatee's oval-shape for counting estimation.

\subsubsection*{Anisotropic Gaussian Kernel for Manatee Counting}
While the oval-shaped density map in Figure~\ref{fig:label_examples}(IV) is resembles to manatees' shape and orientation, it does not effectively capture the width of manatees' body, where the length of a manatee is much longer than its width. AGK kernel is a modification of the Gaussian kernel \eqref{gk}. Instead of using a single kernel parameter for all features, AGK kernel uses different kernel parameters to differentiate Gaussian distributions along each feature dimension. The AGK kernel function can be defined as
\begin{equation}
    \label{agk}
\textbf{K}_{\sigma}(\mathbf{x}_i,\mathbf{x}_j)=\text{exp}(-\sum_{k=1}^{d}\frac{\left\| x_{i,k}-x_{j,k}\right\|^{2}}{2\sigma_k^{2}})
\end{equation}
where $k=1,\cdots,d$ denotes the feature indices. Because manatee's density maps are 2-D images, we have $d=2$ for manatee counting. To generate a density map that most closely resembles to each individual manatee, parameter $\sigma_k$ is dynamically adjusted using Eq.~(\ref{equ:sigmas}).
\begin{equation}
    \begin{cases}
        \sigma_1 =& \frac{\left\| \mathbf{x}_a -\mathbf{x}_b\right\|}{2}\times \frac{\text{FWHM}}{\alpha}\\
        \sigma_2 =& \frac{\sigma_1}{\text{\AR}} 
    \end{cases}
    \label{equ:sigmas}
\end{equation}
in Eq.~(\ref{equ:sigmas}), $\left\| \mathbf{x}_a -\mathbf{x}_b\right\|$ denotes the length of the underlying line-segment. $\text{FWHM}$ denotes Full Width at Half Maximum of the Gaussian distribution defined by the line-segment. It is used to adjust the $\sigma$ to make the distribution more centralized. Intuitively, given a Gaussian distribution with $\sigma$ standard deviation value,  $\text{FWHM}$ and $\sigma$ satisfy following relationship: $\text{FWHM}=2\sqrt{2\ln2}\sigma\approx 2.355\sigma$. To allow flexibility, we use a parameter $\alpha$ to penalize the $\text{FWHM}$, and control Gaussian distribution spread.

Overall, $\sigma_1$ is adjusted based on the length of the line-segment and the $\text{FWHM}$. For $\sigma_2$, it is also adjusted based on the $\sigma_1$ value and the empirical Aspect Ratio (\AR), \textit{i.e.} length divided by width, of the underlying object. In our experiments, we empirically set a fixed \AR value for all experiments. 

\begin{figure*}[ht]
  \centering
  \includegraphics[width=1\textwidth]{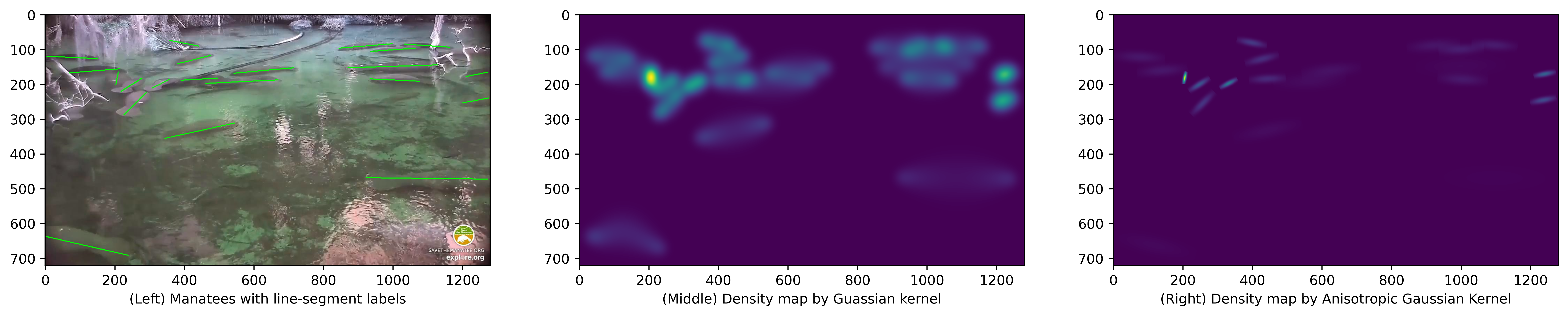}
    \caption{\emph{Left:} An image with line labels of manatees, \emph{Middle:} Density map of the image generated by using generic Gaussian kernels (line-segment labels), and \emph{Right:} Density map generated by using AGK kernels (line-segment labels)}
    \label{fig:line_example}
\end{figure*}

Figure~\ref{fig:label_examples}(V) shows an example of the density map generated for a ling between $(5, 5)$ and $(25, 25)$ by setting the AGK parameters with $\sigma_{1}=8.24$ and $\sigma_{2}=2.06$. Comparing Figure~\ref{fig:point_example}(IV) and (V), \textit{i.e.} density maps generated using simple Gaussian kernel \textit{vs.} Anisotropic Gaussian kernel (AGK) respectively, the results show that AGK based density map is more resemblance to the manatee shape, and therefore will result in more accurate counting for manatees. 

In Figure~\ref{fig:line_example}, we report a manatee aggregation with line-segment labels (left panel), the density map generated from Gaussian kernels (middle), and the density map generated from AGK kernels, by adding oval shaped Gaussian distributions over the lines. Comparing two density maps, we can find that density map from AGK kernels provide more accurate density estimation with respect to individual manatee's shape and position. For example, the two manatees to the upper-right scene are represented as two circular-shaped areas, whereas AGK kernel's density shows clear oval-shapes with accurate orientation and aspect ratios. 

\begin{figure*}[ht]
  \centering
  \includegraphics[width=\textwidth]{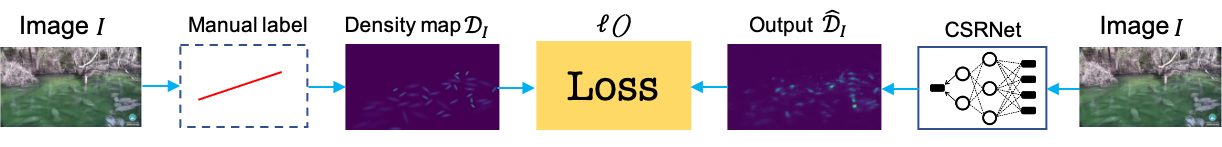}
    \caption{A conceptual view of the deep neural network based manatee counting workflow. From left to middle: An input image ($I$), the line-segment labels of the image, the density map $\mathcal{D}_I$ of the image. From right to middle: An image $I$ is applied to a neural network (NN) to predict its density map $\hat{\mathcal{D}}_I$. The loss function in the middle is used to regulate and train the NN to ensure that $\hat{\mathcal{D}}_I$ maximally approximates to the ground-truth $\mathcal{D}_I$.}
    \label{fig:manatee_framework}
\end{figure*}

\subsection*{CSRNet Learning to Predict Density Map}
After obtaining density maps for each labeled images, our next step is to use original images to train a deep neural network. Once the network is sufficiently trained, for each input image, the network will output a density map maximally approximating to the ground-truth density map of the input image. 

More specifically, given a labelled training deadset $\mathcal{I}$ with $|\mathcal{I}|$ images where each image $I$ is labeled and has a ground-truth density map $\mathcal{D}_I$. Denote $f_{\Theta}()$ a deep neural network regulated by a number of tunable parameters $\Theta$, and for each labeled input image $I_i\in\mathcal{I}$ and its density map $\mathcal{D}_{I_i}\in\mathcal{D}$, the network will output a predicted density map $\hat{\mathcal{D}}_{I_i}$, the neural network aims to learn optimal parameters $\Theta^*$, such that total loss $\ell()$ with respect to the predicted density map is minimized.
\begin{equation}
\Theta^*=\underset{\Theta}{\arg\min}~\sum_{I_i\in\mathcal{I}}\ell(f_{\Theta}(I),\mathcal{D}_{I_i}) 
\label{eq:nnloss}
\end{equation}

To learn to predict density map, we employ Congested Scene Recognition network (CSRNet) as our basic model. CSRNet is designed to understand highly congested scenes and perform accurate count estimation as well as produce high-quality density maps\cite{li2018csrnet}. 
The CSRNet consists of two major components, including a CNN (convolutional neural network) as the front-end for 2-D feature extraction and a dilated CNN for the back-end to replace pooling operations. In our design, we use VGG16~\cite{simonyan2014very}, a pretrained CNN network, as the front-end of CSRNet because it has strong transfer learning ability and flexibility to concatenate the back-end for density map generation. The fully-connected layers of VGG-16 is removed in the CSRNet.

For the back-end, a 2-D dilated convolution can be defined as
\begin{equation}
y(m,n)=\sum_{i=1}^{M}\sum_{j=1}^{N}x(m+r\times i,n+r\times j)w(i,j) 
\label{dilated cnn}
\end{equation}
where $y(m,n)$ is the output, $x(m + r \times i,n + r\times j)$ is the input, $w(i,j)$ is the filter with $M$ length and $N$ width respectively, and $r$ is the dilation rate. When $r=1$, the dilation convolution layer is the same as a normal convolution layer. 

In the dilated convolution layer, a $k\times k$ kernel filter will be enlarged to $(k+(k-1)(r-1))\times (k+(k-1)(r-1))$ with dilated stride $r$. Because of the application of  the dilated convolution, the CSRNet can maintain the resolution of feature map.  Most importantly, the output from dilated
convolution contains more detailed information.

\subsection*{Overall Framework}
Algorithm~\ref{alg_main} outlines major steps of the proposed framework for manatee counting using DNN and AGK kernels. In addition, Figure~\ref{fig:manatee_framework} also shows conceptual flow of the framework with respect to input images, annotations, density maps, and DNN network training. 

The general workflow can be splitted into two phrases, annotation and training phrase, and prediction phrase. During annotation and training phrase, all training images are labelled, and each original image and its labels are used to generate AGK density map (detailed in Algorithm \ref{alg_agk}). After this step, all training images and their density maps are used to train a deep neural network, by using Eq.~(\ref{equ:loss}) to calculate the loss between the DNN output and each training image ground-truth density map. The loss is used to update weights of the DNN until the network converges. After that, it enters into prediction phrase that we can use the trained NN to predict the number of manatees in a test image.

\subsubsection*{Density Map Generation}
One of the main steps in Algorithm~\ref{alg_main} is to generate AGK density map (Algorithm~\ref{alg_agk}) for each image $I$ (this step can also be replaced by Algorithm~\ref{alg_line} to generate normal Gaussian Kernel Density Map). For Gaussian distributions, the probability density values are continuous infinitely with respect to the input space. In reality, the distributions are restricted to a 2-D window, limited by the image size. From distribution perspective, for a generated Gaussian distribution, the sum all values within the 2-D array should be equal to 1. Therefore, we truncate each single Gaussian distribution to a specific region, and all distributions' output, and further normalize the values to ensure that they follow a distribution. In the following, we explain steps to generate three types of density map, dot-label, line-label, and line-label Anisotropic Gaussian, respectively. 

\paragraph{Dot-label Gaussian Density Map} To obtain a dot-label Gaussian density map, $n$ Gaussian distributions are generated based on the number of dots (\textit{i.e.} labels) within the image. Then these $n$ 2D arrays are added to their corresponding point positions in the map, which is an image-sized initialized 2D array without the color dimension.

\paragraph{Line-label Gaussian density map} To get a line-label Gaussian density map, its general algorithm is shown as Algorithm~\ref{alg_line}. Firstly, a $w$ by $h$ shape 2D array, $\mathcal{O}$, is initialized with 0 where $w$ and $h$ are the with and height of the Image $I$, respectively. Then get each $\mathbf{l}$ from labels, $L$, where the label contains two points, start point and end point of the line. A series of points is generated pixel by pixel from start point to end point. For each of the point, a $\sigma$ value is created according to the Eq.~(\ref{equ:line_scale}) followed by creating its corresponding window size. After that, a Gaussian distribution array can be generated and added to a temporary density map $o$ at its position. Once Gaussian distributions are generated for all of points of the line-segment, the $o$ is normalized because one line label is only regarded as one count. Finally, the $o$ is added back to the output $\mathcal{O}$. 

\paragraph{Line-label Anisotropic Gaussian Density Map} Algorithm~\ref{alg_agk} lists steps to generate a line-label anisotropic Gaussian density map. An array of $\mathcal{O}$ is initialized with 0 which is the same as it in generating a line-label Gaussian density map. For each of the label, $\mathbf{l}$, in labels, $L$, first, the length, $len$, of the line-segment is calculated. With the length of the segment, $\sigma_1$ is calculated according to Eq.~(\ref{equ:sigmas}) followed by $\sigma_2$. Then, generating anisotropic Gaussian distribution based on $\sigma_1$ and $\sigma_2$ within a 2-D array, $\mathcal{N}(\mathbf{\mu},\mathbf{\sigma})$, with the shape of $len$ by $len$. After that, it rotates the generated distribution respecting to the slope of the label. Before adding $\mathcal{N}^{\circ}(\mathbf{\mu},\mathbf{\sigma})$ back to the output $\mathcal{O}$ at the position of the center of the line-segment, $\mathcal{N}^{\circ}(\mathbf{\mu},\mathbf{\sigma})$ is normalized. Once all the distributions have been created for the labels, anisotropic Gaussian density map is presented for the image $I$.

\subsection*{{Limitations}}
The proposed anisotropic Gaussian kernel relies Gaussian distributions to generate a density map to approximate manatee shapes. By doing so, it relies on assumption that manatee shapes are largely visible and are close to the AGK kernel shape. As a result, several potential limitations may hinder this approach from obtaining high accuracy counting results:
\begin{itemize}
\item{Close-up Views}: When manatees are situated too close to the camera, they may appear disproportionately large, with only parts of their bodies being captured. In such cases, line-labels fail to accurately mirror the shape of the manatee. The resultant density graph might not correspond well to the actual manatee shape, and result in poor counting performance.

\item{Distant Views}: On the opposite end, manatees that are extremely further away from the camera appear diminutive, resembling dots. Under these conditions, the differentiation between line-label and dot-label becomes negligible, resulting in comparable accuracy for both annotation methods.
\end{itemize}
These real-world scenarios outline the importance of understanding the inherent constraints of different labeling/annotation approaches for real-world applications. Our experiments and comparisons demonstrate the strength and niche of AGK kernels in counting manatee under different environment conditions.

\begin{algorithm}[!ht]
\begin{small}
    \KwData{(1) $\mathcal{I}$: training image set;\\
~~~~~~~~~~  (2) $\mathcal{T}$: test image set;
    }
\textbf{Define:} (1) $\mathcal{D}$: density map set; \\
~~~~~~~~~~~~~~(2) $f_{\Theta}()$: a DNN, \textit{e.g.} a CSRNet or other DNNs; \\
~~~~~~~~~~~~~~(3) $a(I)$: label an image $I$ and return annotations;
\texttt{\\ === Annotation and training phase} ===\\
$\mathcal{D} \leftarrow \{\}$ \\
\For{$I \in \mathcal{I}$} { 
    $\mathbf{L} \leftarrow  a(I)$ \tcp*{obtaining training image $I$ annotations}
    $\mathbf{D} \leftarrow  AGKDensity(I, \mathbf{L})$     \tcp*{obtaining $I$'s AGK density map}
    $\mathcal{D} \leftarrow \mathcal{D} \cup \mathbf{D}$
}
$f_{\Theta^*}() \leftarrow \underset{\Theta}{\arg\min}~\ell(f_{\Theta}(\mathcal{I}),\mathcal{D})$ \tcp*{training DNN network using Eq.~(\ref{eq:nnloss})}

\texttt{\\ === Prediction phase} ===\\
\For{$T \in \mathcal{T}$} { 
    $\hat{\mathcal{D}}^T \leftarrow f_{\Theta^*}(T)$ \tcp*{predicting density map}
    $C^T \leftarrow$ Calculate manatee numbers in $T$ using Eq.~(\ref{equ:counting})
}

\caption{DNNs for Manatee Counting Using AGK}
\label{alg_main}
\end{small}
\end{algorithm}

\begin{algorithm}[!ht]
\begin{small}
  \SetAlgoLined
    \KwData{(1) $I$: an input image; \\ 
~~~~~~~~~~  (2) $L$: line labels of image $I$}
\textbf{Define:} (1) $\mathbf{l}_s$ and $\mathbf{l}_s$: two endpoints of line segment $\mathbf{l}$\\
~~~~~~~~~~~~~(2) $\mathcal{N}(\mathbf{\mu},\mathbf{\sigma})$: a multidimensional Gaussian distribution with $\mathbf{\mu}$ mean and $\mathbf{\sigma}$ spreads \\
\textbf{Input:} (1) $a$: expanding factor; \\
~~~~~~~~~~~~(2) $\sigma_{basic}$: basic value for generating $\sigma$; \\
\textbf{Output:} $\mathcal{O}$: density map of image $I$

$\lbrack w, h \rbrack \leftarrow$ Width and height of image $I$ 
${\mathcal{O}\in\mathbb{R}^{w\times h}} \leftarrow \mathbf{0}$ \tcp*{create an empty density map}
\For{$\mathbf{l} \in L$} {
    $\lbrack\mathbf{x}_a, \mathbf{x}_b\rbrack \leftarrow \lbrack\mathbf{l}_s, \mathbf{l}_e\rbrack$ \tcp*{two ends of a line-segment} 
    $\mathbf{P} \leftarrow$ Evenly select $\lceil |\mathbf{x}_{a.x} - \mathbf{x}_{b.x}| + 1\rceil$ points from line-segment $\mathbf{l}$\\
    ${\circ \in\mathbb{R}^{w\times h}} \leftarrow \mathbf{0}$ \tcp*{temporary density map}
    \For{$\mathbf{p} \in \mathbf{P}$} {
        $\mu \leftarrow \mathbf{p}$ \tcp{Gaussian distribution mean}
        $\sigma \leftarrow \sigma_{basic}+a \cdot \min(\left\|\mathbf{p}- \mathbf{x}_a\right\|, \left\|\mathbf{p} - \mathbf{x}_b\right\|)$ \tcp*{Gaussian distribution spread Eq.~(\ref{equ:line_scale})}
        $\mathcal{N}(\mathbf{\mu},\mathbf{\sigma}) \leftarrow $ Generate 2\-D Gaussian Distribution \\
        $\circ \leftarrow \circ + \mathcal{N}(\mathbf{\mu},\mathbf{\sigma})$ \tcp*{add current Gaussian distribution to the temporary density map}        
    }
    Normalize temporary density map $\circ$ \\
    $\mathcal{O} \leftarrow \mathcal{O} + \circ$
}
\Return $\mathcal{O}$

\caption{LineDensity() - Gaussian Kernel Density Map Generation via Line Labels}
\label{alg_line}
\end{small}
\end{algorithm}


\begin{algorithm}[!ht]
\begin{small}
  \SetAlgoLined
    \KwData{(1) $I$: an input image; \\ 
~~~~~~~~(2) $L$: line labels of image $I$}

\textbf{Define:} (1) $\mathbf{l}_s$ and $\mathbf{l}_s$: two endpoints of line segment $\mathbf{l}$\\
~~~~~~~~~~(2) $\mathcal{N}(\mathbf{\mu},\mathbf{\sigma})$: a multidimensional Gaussian distribution with $\mathbf{\mu}$ mean and $\mathbf{\sigma}$ spreads

\textbf{Output:} $\mathcal{O}$: density map of image $I$.
$\lbrack w, h \rbrack \leftarrow$ Width and height of image $I$ \\
${\mathcal{O}\in\mathbb{R}^{w\times h}} \leftarrow \mathbf{0}$ \tcp*{create an empty density map}
\For{$\mathbf{l} \in L$} {
    $\lbrack\mathbf{x}_a, \mathbf{x}_b\rbrack \leftarrow \lbrack\mathbf{l}_s, \mathbf{l}_e\rbrack$ \tcp*{two ends of a line-segment}
    $\mathbf{\mu}=\lbrack\mu_1, \mu_2\rbrack\leftarrow \frac{\mathbf{x}_a+\mathbf{x}_b}{2}$ \tcp*{calculate AGK mean}
    $\mathbf{\sigma}=\lbrack\sigma_1, \sigma_2\rbrack\leftarrow$ Eq.~(\ref{equ:sigmas}) \tcp*{calculate AGK spreads}
    $\mathcal{N}(\mathbf{\mu},\mathbf{\sigma}) \leftarrow$ Generate 2-D Gaussian Distribution \\    
    ${\circ} \leftarrow $ Calculate slope between $\mathbf{x}_a$ and $\mathbf{x}_b$ \\
    $\mathcal{N}^{\circ}(\mathbf{\mu},\mathbf{\sigma}) \leftarrow$ Align 2-D Gaussian Distribution $\mathcal{N}(\mathbf{\mu},\mathbf{\sigma})$ based on the line-segment slope\\    
    Normalize $\mathcal{N}^{\circ}(\mathbf{\mu},\mathbf{\sigma})$ \\
    $\mathcal{O} \leftarrow \mathcal{O} + \mathcal{N}^{\circ}(\mathbf{\mu},\mathbf{\sigma})$ \tcp{add current Gaussian distribution to the density map}
}
\Return $\mathcal{O}$

\caption{AGKDensity() - AGK Density Map Generation via Line Labels}
\label{alg_agk}
\end{small}
\end{algorithm}

\section*{Experiments}

\subsection*{Benchmark Data}

\vspace{0.1cm}\noindent

In order to validate the performance of proposed framework using low resolution images, manatee surveillance video clips from ``Save the manatee Club''~\footnote{https://www.savethemanatee.org/manatees/manatee-webcams/}, which are captured from webcams placed at Blue Spring State Park, are collected to create our benchmark dataset. 
After using FFmpeg to generate images from the video clips and removing similar images, we obtain 784 images, consisting of different number of manatees, as our testbed.

\subsubsection*{Deduplication}
Because manatee images are collected from video clips and manatees rarely show rapid movement, some of the images are similar to each other. To avoid data duplication, duplicate images are dropped 
by using VGG-16 to extract layer-wise features for each image and then a method is employed to calculate the difference between two images, $I_a$ and $I_b$, with the formula being defined as follows:
\begin{equation}
    \text{Disance}(I_a, I_b) = \sum_{j \in \mathbb{F}}\frac{1}{C_{j}H_{j}W_{j}}\left||\psi_{j}(I_a) - \psi_{j}(I_b) \right||_{2}^{2}
    \label{equ:similarity}
\end{equation}
$\mathbb{F}$ is the outputs from the last activation layer in each group of VGG-16's 5 groups, namely $\mathbb{F} = \{\psi_{j}|j=1,2,3,4,5\} = \{layer_2, layer_4, layer_7, layer_{10}, layer_{13}\}$. ${C_{j}, H_{j}, W_{j}}$ are the three dimensions of the image, channel, height, and width, respectively. $\psi_{j}(i_x)$ denotes the $j$th group ReLu output of the image $i_x$. The smaller value of $\text{Distance}(I_a, I_b)$, the more similarity of the two images is considered. When the value of $\text{Distance}(I_a, I_b)$ is less than 2, the two images are considered highly similar, then one of the two images is dropped. 

After comparing similarities of the images, 415 of 784 images are discarded and only 369 images are kept as dataset.

\subsubsection*{Dataset Characteristics}
All 369 benchmark images have the same size. According to the number of manatees in each image, we separate each image into three density levels: low, medium, and high. We define that, if an image has less than 5 manatees, then it is of low density level. Continue with medium level if it has more or equal to 5 but less than 20 manatees. If an image with more than 19 manatees, it is of high density level.

\begin{figure}[ht]
    \centering
    \includegraphics[width=0.35\textwidth]{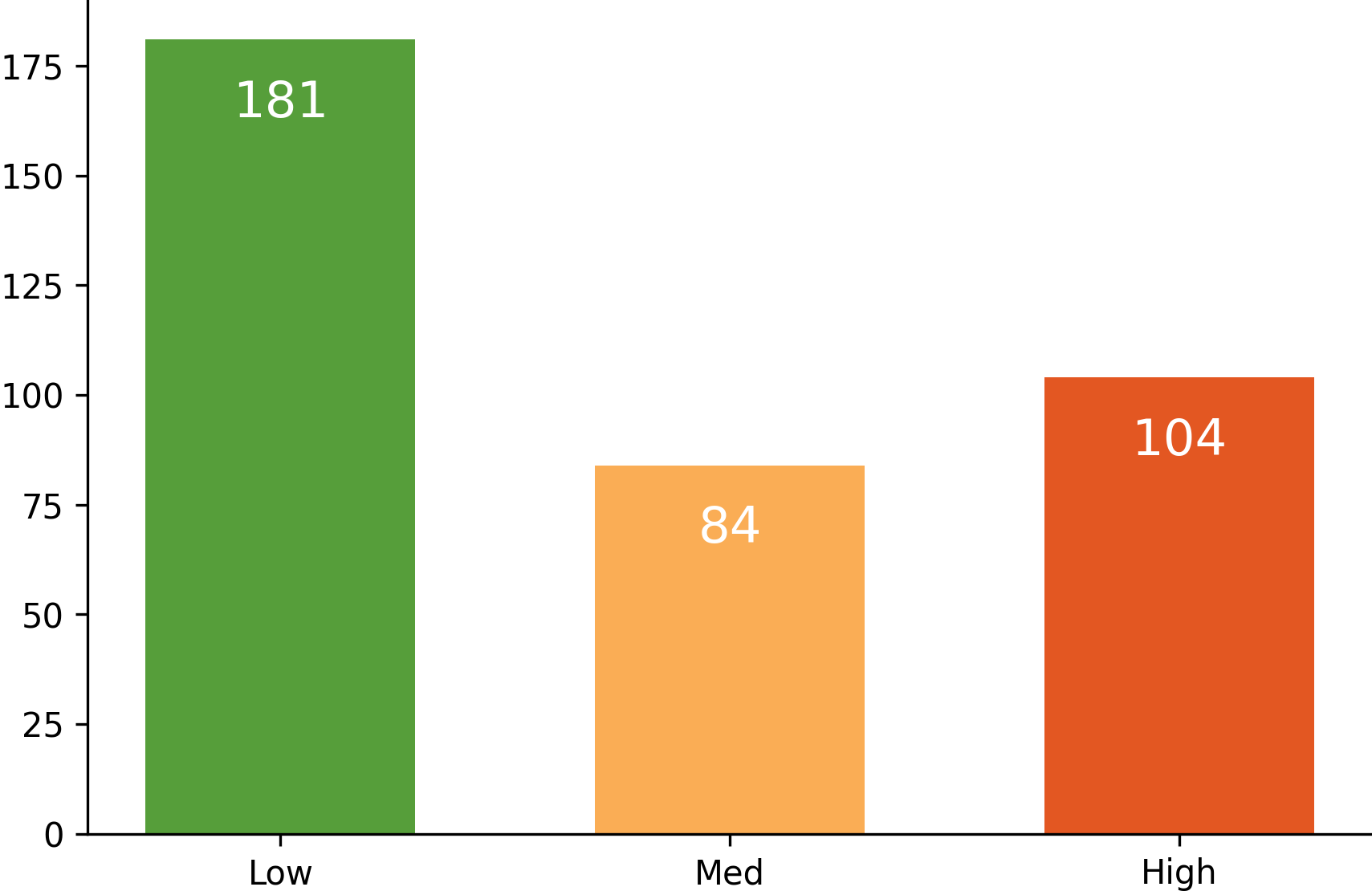}
    \caption{Distribution of benchmark images \textit{w.r.t.} different manatee density levels. The $x$-axis denotes manatee density levels and the $y$-axis denotes number of images.}
    \label{fig:manatee_density_levels}
\end{figure}

Figure~\ref{fig:manatee_density_levels} reports the manatee density distributions in the benchmark images at three levels. Images with low density level have the highest quantity (181), which is almost equal to the total amount of medium and high level which have 84 and 104 images, respectively. 

\begin{figure*}[ht]
    \centering
    \includegraphics[width=1\textwidth]{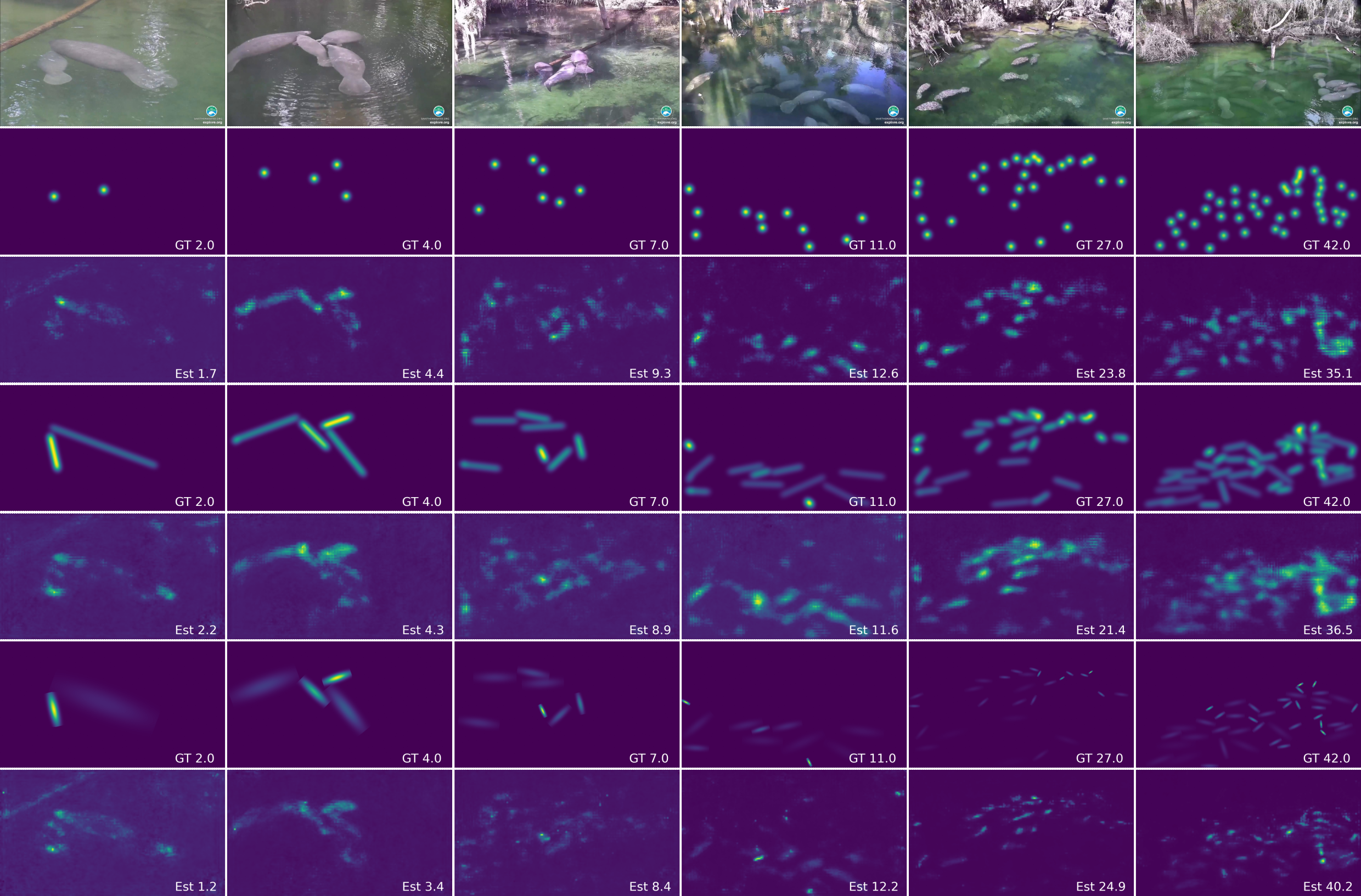}
    \caption{Examples of algorithm performance with respect to different manatee densities in the scene. The first row shows original images with increasing manatee density from left to right. The second and third rows show ground-truth density map (2nd row) and predicted density map (3rd row) using dot labels. The fourth and fifth rows show ground-truth density map (4th row) and predicted density map (5th row) using line labels and generic Gaussian kernels. The sixth and seventh rows show ground-truth density map (6th row) and predicted density map (7th row) using line labels and anisotropic Gaussian kernels}
    \label{fig:density_levels}
\end{figure*}

\begin{table*}[ht]
    \centering
    \begin{small}
    \tabcolsep 12pt
    \caption{\small{Manatee counting performance using different DNNs over different density levels and label types}}
    \begin{tabular}{l l l l l l l l l }
    \hline\hline
        \multirow{2}{*}{Method} & \multicolumn{2}{ c }{Low} & \multicolumn{2}{ c }{Medium} & \multicolumn{2}{ c }{High} & \multicolumn{2}{ c }{Overall} \\ \cline{2-9}
        & MAE & RMSE & MAE & RMSE & MAE & RMSE & MAE & RMSE  \\ \hline
        MCNN-dot          & 3.640 & 3.812 & \textbf{3.511} & 4.924 & \textbf{26.982} & \textbf{28.352} & \textbf{11.377} & 12.363  \\ 
        MCNN-line         & 3.603 & 3.753 & 3.556 & \textbf{4.912}  & 26.998  & 28.375 & 11.386 & 12.346  \\ 
        MCNN-anisotropy   & \textbf{2.471} & \textbf{2.859}  & 4.170  & 5.562  & 28.080 & 29.408 & 11.573 & 12.610  \\ \hline\hline
        SANet-dot          & 2.048 & 3.032 & 2.833 & 3.909 & 4.993 & 7.510 & 3.291 & 4.817  \\ 
        SANet-line         & \textbf{1.512} & \textbf{2.645} & \textbf{2.268} & \textbf{3.418}  & \textbf{4.197} & \textbf{6.882} & \textbf{2.659} & \textbf{4.315}  \\ 
        SANet-anisotropy   & 1.847 & 2.619  & 3.876  & 5.047 & 6.042 & 8.586 & 3.921 & 5.417  \\ \hline\hline
        VGG-dot          & 3.921 & 4.721 & 2.702 & 3.522 & 15.601 & 17.199 & 7.408 & 8.481  \\ 
        VGG-line         & 3.723 & 4.540 & 2.678 & 3.535 & 16.217 & 17.748 & 7.539 & 8.608  \\ 
        VGG-anisotropy   & \textbf{3.405} & \textbf{4.125} & \textbf{2.550} & \textbf{3.488} & \textbf{15.311} & \textbf{16.888} & \textbf{7.088} & \textbf{8.167}  \\ \hline\hline
        MARUNet-dot & \textbf{2.312} & \textbf{2.738} & 4.292 & 5.685 & 28.150 & 29.447 & 10.045 & 15.982  \\ 
        MARUNet-line & 2.653 & 3.578 & 4.347 & 5.797  & 27.046  & 28.498 & 9.913 & 15.583  \\ 
        MARUNet-anisotropy & 3.325 & 4.122  & \textbf{3.830}  & \textbf{5.030} & \textbf{25.063} & \textbf{26.677} & \textbf{9.567} & \textbf{14.651}  \\ \hline\hline
        
        CSRNet-dot          & \textbf{1.344} & \textbf{1.925} & 2.981 & \textbf{3.825} & 4.978 & 6.569 & 3.101 & 4.106  \\ 
        CSRNet-line         & 2.256 & 3.341 & 3.669 & 4.67  & 5.03  & 6.614 & 3.652 & 4.875  \\ 
        CSRNet-anisotropy   & 1.588 & 2.19  & \textbf{2.94}  & 3.84  & \textbf{4.506} & \textbf{5.856} & \textbf{3.011} & \textbf{3.962}  \\ \hline\hline
    \end{tabular}
    \label{tab1}
    \end{small}
\end{table*}
\subsection*{Experimental Settings}
\subsubsection*{Manatee Annotations}
To validate the performance of using Anisotropic Gaussian Kernel (AGK) for manatee counting, we use dot and line labels to annotate images for performance comparison. In order to reduce duplicate manual label work, we draw a line over each manatee (from tail to head) and make sure the line crosses over the center point of the manatee (the number of labelled line segments within an image is the number of manatees in the image). Consequently, two endpoints of a line for each manatee are obtained and saved into a JSON file for further usage. Assuming $\mathbf{x}_a=\lbrack x_{a,x}, x_{a,y}\rbrack$ and $\mathbf{x}_b=\lbrack x_{b,x}, x_{b,y}\rbrack$ are the two endpoints of a labelled line segment obtained from previous stage, $\lbrack \frac{x_{a,x} + x_{b,x}}{2}, \frac{x_{a,y} + x_{b,y}}{2} \rbrack$ is used as point label for the same manatee. 

\subsubsection*{Parameter Settings}
We validate the performance of multiple baseline methods on our manatee dataset over three labels. Compared to traditional point labeling methods~\cite{li2018csrnet}, our line labeling adds only a small amount of additional time and obtains an overall better performance. In this section, we report the evaluation metrics and perform an ablation study on the three labeled datasets to study their performance.

We evaluate the object counting task on an NVIDIA V100 GPU card. The watermark on the surveillance video images at the lower right corner is filled with solid black to avoid additional interference. During training, the initial learning rate is 1e-4, and the Adam optimizer is used. For better training and to prevent overfitting, random flips were used for augmentation. For all networks, the batch size is set to 4.

The network directly converts the input image into a density map in training. During the density map generations, some hyperparameters are used in Algorithms~\ref{alg_line} and~\ref{alg_agk}. They are the expanding factor, $a=0.2$, the base value of $\sigma$, $\sigma_{basic}=15$, the empirical Aspect Ratio, $\AR=4$, full width at half maximum \text{FWHM} = 2.355$\approx2\sqrt{2\ln2}$, and the \text{FWHM} penalizer $\alpha$ in Eq.~(\ref{equ:sigmas}) is set as $\alpha=4$. Intuitively, the \text{FWHM} and $\alpha$ parameters in Eq.~(\ref{equ:sigmas}) are determined such that the $\sigma_1$ value is roughly $\frac{1}{4}$ of the length of the underlying line-segment $||\mathbf{x}_a - \mathbf{x}_b||$.

Same as in previous studies \cite{li2018csrnet}, we chose the MSE loss to measure pixel difference between ground truth and predicted density maps, with loss value being calculated as follows:
\begin{equation}
    \begin{split}
        \ell() = \frac{1}{|\mathcal{T}|}\sum^{|\mathcal{T}|}_{i=1}||\mathcal{D}_{T_i}-\hat{\mathcal{D}}_{T_i}||^2_2
    \end{split}
    \label{equ:loss}
\end{equation}
where $\mathcal{T}$ denotes the test set and $|\mathcal{T}|$ denotes number of images in the test set. For each test image $T_i$, $\mathcal{D}_{T_i}$ and $\hat{\mathcal{D}}_{T_i}$ denote ground-truth and predicted density maps, respectively. $||\cdot||^2_2$ represents the Euclidean distance.

\subsection*{Performance Metrics}
We use mean absolute error (MAE) and root mean square error (RMSE) in the experiments to evaluate algorithm performance.
\begin{equation}
    \begin{split}
        MAE = \frac{1}{|\mathcal{T}|}\sum^{|\mathcal{T}|}_{i=1}|C_{T_i}-\hat{C}_{T_i}|
    \end{split}
    \label{equ:mae}  
\end{equation}
\begin{equation}
    \begin{split}
        RMSE = \sqrt{\frac{1}{|\mathcal{T}|}\sum^{|\mathcal{T}|}_{i=1}(C_{T_i}-\hat{C}_{T_i})^2}
    \end{split}
    \label{equ:rmse}
\end{equation}
where $C_{T_i}$ and $\hat{C}_{T_i}$ are the ground-truth number and predicted number of manatees in image $T_i$, respectively, calculated using Eq.~(\ref{equ:counting}). 

\subsection*{{Baselines}}
MCNN\cite{zhang2016single}, SANet\cite{cao2018scale}, VGG\cite{simonyan2014very}, and CSRNet\cite{li2018csrnet} are some of the commonly used networks for counting problem. Meanwhile, attention mechanisms were also adopted for counting task recently, such as MARUNet\cite{rong2021coarse}. 
Those models are implemented and trained for comparisons to demonstrate the effectiveness of our proposed method. In the following, we use X-dot, X-line, and X-anisotropy to denote that X neural network is trained by using three different types of density maps, which are generated by point label with Gaussian kernel, line label with Gaussian kernel, and line label with anisotropy kernel, respectively.
\begin{itemize}

\item{MCNN\cite{zhang2016single}:} MCNN employs a multi-column CNN structure that is adaptive to detect heads of varying sizes, addressing scale variations inherent in crowd images and generating density maps for crowd estimation.
\item{SANet\cite{cao2018scale}:} The Scale Aggregation Network employs a feature map encoder to aggregate multi-scale features from original images. Then it uses a density map estimator to fuse these features before generating high-resolution density maps. This architecture ensures robust crowd counting across varied crowd densities.
\item{VGG\cite{simonyan2014very}:} The VGG network, characterized by its depth and 3x3 convolutions, is known for its robust feature extraction capabilities and has been a foundational architecture in various visual tasks, including crowd counting. 
\item{CSRNet\cite{li2018csrnet}:} CSRNet uses the simplified VGG-16 as front-end structure for feature extraction and a dilated convolution network as back-end, which can handle the scale variations in crowd counting tasks effectively.
\item{MARUNet\cite{rong2021coarse}:} MARUNet, Multi-level Attention Refined UNet, integrates a density map estimator and a crowd region recognizer, facilitating the network's focus on crowd regions and providing a strong baseline in crowd density map generation. 
\end{itemize}

\subsection*{Experimental Results}
Figure~\ref{fig:density_levels} reports two examples for each level of the three density levels, low, medium and high with three different density maps. The first row are the original images that each image has 2, 4, 7, 11, 27 and 42 manatees separately. The second row is their Gaussian kernel density maps generated from point labels and the fourth row is their Gaussian kernel density maps generated by line labels while the sixth row is their anisotropic Gaussian kernel density maps. And third, fifth and seventh row are their corresponding estimated results from trained models. The number at the bottom right corner on the each of the density map is the real or estimated amount of manatee in the image.

When comparing density maps from the training results with their original counterparts, as shown in Figure~\ref{fig:density_levels}, it shows that as the number of manatees in the image increases, AGK density maps can represent their original maps more closely compared to the other two density maps. The reason is that in high-density maps, manatees appear relatively smaller and possess similar sizes. As a result, the AGK-generated density provides a more accurate representation of their shapes with the given experimental hyper-parameters. In contrast, the density maps generated from GK-point and GK-line tend to distribute more widely rather than concentrating in specific manatees' areas.

For low and medium density maps, manatees tend  to appear larger within the images. In extreme cases, a single manatee might occupy more than one-third of the entire image with only part of its body appearing in the image. Under such conditions, all three types of density maps: AGK, GK-dot, and GK-line, struggle to represent the manatees accurately, resulting in deteriorated performance. Interestingly, in these specific scenarios, GK-dot and GK-line potentially outperform AGK. This is because of their inherent capability to represent larger objects more effectively. For AGK, constrained by preset hyper-parameters, it is more adept at depicting crowded manatees, which are often of smaller sizes.

The results in Table \ref{tab1} show that the anisotropy method has the lowest MAE and RMSE values across the VGG, MARUNet, and CSRNet models for the overall dataset. This shows the efficacy of our proposed line-label anisotropic Gaussian density map. Although this method does not surpass the other two when applied to the MCNN model, the performance metrics are remarkably close among all three methods. For the SANet model, the line Gaussian method yielded the most favorable results.

Among all five DNNs, SANet and CSRNet have better performance than others especially in high density maps, where MCNNs and MARUNets have extremely high MAE and RMSE values. The performance of SANet and CSRNet are very similar. In low-density scenarios, CSRNet shows slight advantages, but in medium to high-density scenarios, SANet performs better. Overall, SANet has a lower MAE while CSRNet has better stability. Interestingly, among the four compared networks, SANet is the only one that performed best under the line-label. This may be attributed to its unique scale aggregation architecture. This empirical success proves that our proposed line-label is indeed effective for crowd manatee counting task.

CSRNet-dot has the best MAE and RMSE values in low density cases while it is outperformed by CSRNet-anisotropy in high density and overall cases. Specifically, in the low density map, where manatee localizations are more distinct, CSRNet-dot registers the lowest MAE and RMSE values of 1.344 and 1.925, respectively. When the density increases, the proposed CSRNet-anisotropy shows superior MAE values in medium to high density scenarios compared to both CSRNet-dot and CSRNet-line. The superior performance of CSRNet-dot in low densities can be attributed to the clearer object locations enabled by dot annotations, given that CSRNet can effectively separate localization from counting tasks.

In terms of overall performance in CSRNet, which tested over the all data, our proposed CSRNet-anisotropy method outperforms the other two methods both in the MAE metric and in the RMSE metric, with a MAE value of $3.011$ and a RMSE value of $3.962$. The MAE overall performance of our proposed method is $2.99\%$ better than the CSRNet-dot method and $21.29\%$ better than the CSRNet-line method. And the RMSE overall performance of our proposed method is $3.63\%$ better than the CSRNet-dot method and $23.04\%$ better than the CSRNet-line method.

In some cases, the proposed anisotropy kernel does not consistently outperform other models, especially when combined with DNN networks like MCNN and SANet. MCNN struggles to identify the optimal method across the three different density levels. In contrast, SANet produces its best performance with the line kernel. Such discrepancies could stem from the intrinsic model structures associated with image scaling in MCNN and SANet. It becomes apparent that our method may not be universally effective for all DNN models.

Another noteworthy observation is the subpar performance of our method in low-density map levels across four out of five tested models. This might be attributed to the fact that, in low-density images, manatees often appear larger or may only partially be in the frame. Such scenarios might not be ideally represented by the density maps generated through the anisotropy kernel especially with the given experimental hyperparameters which used for all images. In these instances, Gaussian kernels with dot or line annotations might provide a better fitting representation.

In the subsequent section, to further substantiate the efficacy of our method, we apply the proposed method to wheat head counting, where objects have regular consistent shapes.

\subsection*{{Additional Results: Wheat head counting}}

\begin{figure*}[ht]
  \centering
  \includegraphics[width=\textwidth]{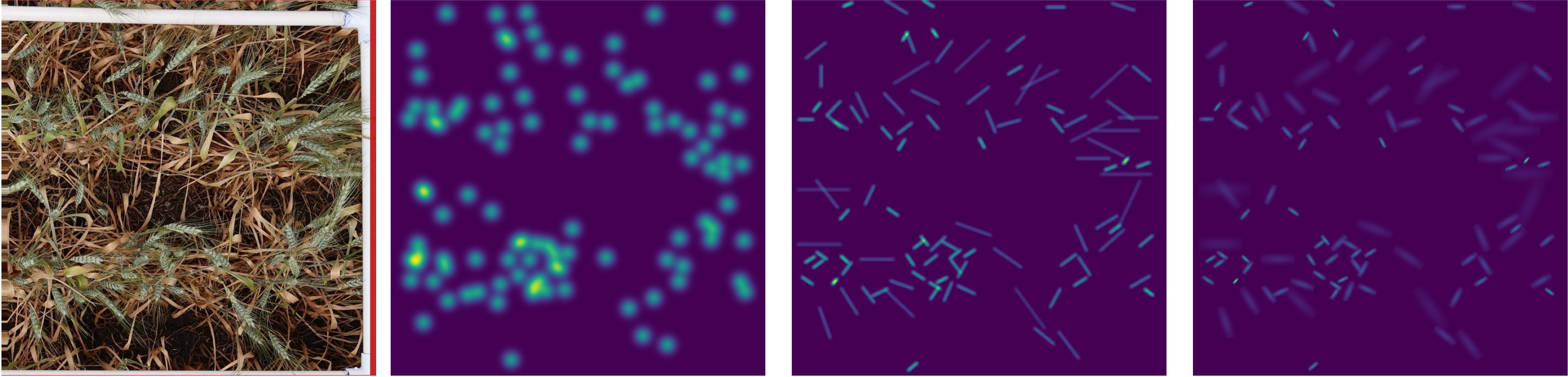}
    \caption{The original wheat image and its three types of corresponding density maps (left to right): dot based GK, line based GK,  and line based AGK}
    \label{fig:wheat_head}
\end{figure*}

In the previous section, we validated our proposed method in the newly created manatee counting dataset. Given the intricate backgrounds and the variability in manatee shapes, often resulting from varying camera distances, the label can not present the manatee over different density levels as shown and analyzed in prior sections.

To further validate the ability and adaptability of our method, we have opted to test it on the "Global Wheat Head Dataset 2021"\cite{david2021global}. Unlike manatees, most wheat heads in these images possess consistent, elongated rectangle-like shapes. We posit that our proposed line annotation with AGK could intuitively represent such forms more effectively.

Figure \ref{fig:wheat_head} shows an example of original wheat head images and three corresponding density maps that are generated by a dot-based Gaussian Kernel(GK), a line-based GK and a line-based anisotropic GK. It is obvious that line-based AGK density map has a better representation of the wheat heads in terms of position, shape, and direction.

\begin{table*}[ht]
    \centering
    \begin{small}
    \tabcolsep 15pt
    \caption{\small{Wheat head counting performance in CSRNet over different label types}}
    \begin{tabular}{l l l}
    \hline\hline
        Method & MAE & RMSE  \\ \hline
        Yolo-v5(GWC-2021)\cite{david2023global} & NA    & 67.0 \\
        CSRNet-dot          & 30.1 & 37.6 \\
        CSRNet-line         & 20.3 & 28.1 \\
        CSRNet-anisotropy   & 15.7 & 20.2  \\ \hline\hline
    \end{tabular}
    \label{tab2}
    \end{small}
\end{table*}

As illustrated in Table \ref{tab2}, Yolo-v5(GWC-2021)\cite{david2023global} has the poorest performance in terms of RMSE. All three models based on CSRNet outperform Yolo-v5, with CSRNet-anisotropy has outstanding MAE of 15.7 and RMSE of 20.2. Yolo-v5 is detection based algorithm that while it has good bounding box accuracy, it misses numerous wheat heads, especially those that are smaller in size. The average size of those missed wheat heads is about 35\% smaller than the average size of all wheat heads.

For our density estimation-based method, instead of seeking the precise location and boundary of each wheat head directly, it determines the total number of wheat heads by estimating the density of wheat heads within a particular region. This method is particularly good at detecting smaller and densely distributed wheat heads since it emphasizes the overall quantity estimation rather than pinpointing individual targets.

When the MAE and RMSE values are close, it often indicates there are not significant outliers. Such consistency can suggest that the model's predictions are relatively accurate, with minimal discrepancies. As shown in Table \ref{tab2}, the proximity of the MAE and RMSE values reinforces this notion of accuracy in the CSRNet's predictions, especially CSRNet-anisotropy which proves the effectiveness of our method.

Throughout the training process, we noticed that the model effectively capture the macro structure of the wheat. For point label, the model had to work hard to fit the density map precisely to a circular region. To address this challenge, we introduced AGK to generate density map by using line labels, which differ significantly from the traditional circular Gaussian region. This label utilizes an elliptical anisotropic Gaussian distribution to present the wheat's location and shape. In this distribution, the core line represents the exact location of the wheat head, while the surrounding decay area outlines the potential uncertainty of the position. Compared to the conventional circular Gaussian region, this method endows the model with enhanced adaptability in wheat head counting.

\section*{Conclusion}

In this paper, we proposed a deep neural network (DNN) based crowd counting approach to count manatee aggregations. This method capitalizes on low-quality images to count manatees in a designated region. Although crowd counting has been used in many other applications (\textit{e.g.} counting cars or audiences), we argued that manatee counting has unique challenges, including surface reflection, occlusions, camouflage. To reduce labelling costs, we employed line-label based annotation, with a single straight line being used to mark each manatee. To take unique shapes of manatees into consideration, we proposed to use Anisotropic Gaussian Kernel (AGK) transform input images into manatee customized density
maps, and then train deep neural networks to learn to count manatee numbers automatically using a predicted density map. Experiments and comparisons, using low resolution real-world manatee images, show that AGK based counting outperforms other baselines, including traditional Gaussian kernel based approach. The proposed approach works particularly well when the image has a high density of manatees in complicated background.

Our findings demonstrate a promising trajectory for broader applications. By transitioning from dot to line labeling, we not only enhanced the accuracy in counting manatees but also successfully improved wheat head counting. The proposed methodology holds potential for various applications, especially for entities with convex-shaped objects, including diverse animals such as livestock like sheep and cattle, and crops such as wheat head, corn, eggplant, \textit{etc.}

In this study, we are primarily focused on images captured from the water surface. Counting manatees in complex underwater backgrounds remains an open problem~\cite{9964179}. 
Future study may take manatees’ movement into consideration to improve counting accuracy, as static objects like branches and rocks remain relatively unchanged over short duration.

\section*{Acknowledgements}
This research is sponsored by the U.S. National Science Foundation through Grant Nos. IIS-1763452 and IIS-2302786.

\section*{Author contributions statement}
Drafting of the manuscript: Z.W., Y.P., C.U. Design and modeling: Z.W., Y.P., C.U., X.Z. Data collection and analysis: Z.W., Y.P., C.U. Obtained funding: X.Z. Supervision: X.Z.

\section*{Data Availability Statement}
Source code and datasets generated and/or analysed in this study are available in the following GitHub repository: https://github.com/yeyimilk/deep-learning-for-manatee-counting.

\bibliography{ref}

\end{document}